%% file: ms.tex
\documentclass[a4paper]{article}

\usepackage{FORMAT}

\usepackage{float}
\usepackage{xcolor}
\usepackage{url}
\usepackage{booktabs}
\usepackage{paralist}
\usepackage{enumitem}
\usepackage[export]{adjustbox}
\usepackage{multirow}

\usepackage[detect-all]{siunitx}
\usepackage{hyperref}
\hypersetup{%
 pdfauthor = {},
  pdftitle = {},
  pdfsubject = {},
  pdfcreator = {LaTeX with hyperref package},
  pdfproducer = {pdflatex}
  bookmarks=false%
,bookmarksnumbered=true%
,hypertexnames=false%
,breaklinks=true%
,colorlinks=true%
,linkcolor=blue%
,citecolor=blue%
,urlcolor=blue%
}

\usepackage{caption}
\input{macros}
\title{Prosody Modifications for Question-Answering in Voice-Only Settings}
\name{%
Aleksandr Chuklin$^1$, %
Aliaksei Severyn$^1$, %
Johanne R.~Trippas$^2$,\\ %
Enrique Alfonseca$^1$, %
Hanna Silen$^3$, %
Damiano Spina$^2$}
\address{%
$^1$ Google -- Z\"urich, Switzerland\\
$^2$ RMIT University -- Melbourne, Australia\\
$^3$ Google -- London, UK}
\email{%
\{chuklin, severyn, ealfonseca, silen\}@google.com\\
\{johanne.trippas, damiano.spina\}@rmit.edu.au
}

\begin{document}
\maketitle

\input{00-abstract}

\input{01-introduction}

\input{02-methodology}

\input{03-setup}

\input{04-results}

\input{05-conclusions}

\bibliographystyle{IEEEtran}
{\raggedright
\bibliography{mybib}
}

\end{document}

%% file: macros.tex
\newcommand{\squishlist}{
 \begin{list}{$\bullet$}
  { \setlength{\itemsep}{0pt}
     \setlength{\parsep}{3pt}
     \setlength{\topsep}{3pt}
     \setlength{\partopsep}{0pt}
     \setlength{\leftmargin}{1.5em}
     \setlength{\labelwidth}{1em}
     \setlength{\labelsep}{0.5em} } }

\newcommand{\squishlisttwo}{
 \begin{list}{$\bullet$}
  { \setlength{\itemsep}{0pt}
     \setlength{\parsep}{0pt}
    \setlength{\topsep}{0pt}
    \setlength{\partopsep}{0pt}
    \setlength{\leftmargin}{2em}
    \setlength{\labelwidth}{1.5em}
    \setlength{\labelsep}{0.5em} } }

\newcommand{\squishend}{
  \end{list}  }

%% file: 00-abstract.tex
\begin{abstract}
Many popular form factors of digital assistants---such as Amazon Echo, Apple
Homepod, or Google Home---enable the user to hold a conversation with these
systems based only on the speech modality. The lack of a screen
presents unique challenges.
To satisfy the information need of a user, %
the presentation of the answer needs to be optimized for such voice-only
interactions.
In this paper, we propose a task of evaluating the usefulness of audio transformations
(i.e., prosodic modifications)
for voice-only question answering.
We introduce a crowdsourcing setup where we evaluate the quality of
our proposed modifications along multiple dimensions
corresponding to the informativeness, naturalness,
and ability of the user to identify key parts of the answer. 
We offer a set of prosodic modifications that highlight
potentially important parts of the answer using various acoustic cues.
Our experiments show that some of these modifications lead
to better comprehension %
at the expense of only slightly degraded naturalness of the audio.
\end{abstract}

%% file: 01-introduction.tex
\section{Introduction}\label{sec:intro}

Recent advances in technology have transformed the ways we access information.
With the rise of voice-only digital assistant devices, such as
Amazon Echo, \footnote{\url{https://www.amazon.com/echo}}
Apple Homepod, \footnote{\url{https://www.apple.com/homepod/}}
or Google Home \footnote{\url{https://home.google.com}}
users can express information needs verbally and receive answers exclusively via voice.
However, providing answers via voice in the absence of a screen is a challenging task which leads to different interaction strategies employed by both users and the system.

Searching is traditionally considered as a visual task since reading information-dense sections such as search snippets is already a cognitively demanding undertaking.
Thus, screen-based systems typically provide visual cues to \emph{highlight} key parts of text responses (e.g., boldfacing key parts in passages) which helps to identify answers while skimming a results page. However, the serial nature of audio-only communication channels hampers ``skimming'' the information as can be done in a visual interface~\cite{love2005understanding}.
Moreover, the processing of spoken information has a major impact on
user satisfaction due to cognitive and memory limitations~\cite{weng2004conversational}.

To overcome these challenges imposed by speech-only interactions it has been suggested that systems should become actively involved in the search process~\cite{trippas2018informing}.
Previous research shows that users tend to express more complex queries and engage in a dialog with the system~\cite{trippas2018informing}, while for the system it appears to be beneficial to summarize and shorten the answers~\cite{filippova2015sentence}.

In this paper, we explore different prosody modifications---such as insertion of pauses, decreasing of speaking rate, and increase in pitch---to highlight key answer parts in audio responses. While these features of prosody in natural speech have been associated with positive effects, to our knowledge they have not been analysed empirically for presenting answers in voice-only channels.  Moreover,  it remains unclear which effects it would have when incorporated in a voice QA system and how these effects can be evaluated at scale.

We propose to address this problem by asking the following research questions:
\begin{enumerate}[label=\textbf{RQ\arabic*},leftmargin=*,topsep=5pt]
    \item Can we use crowdsourcing to quantify the utility of the prosody
      modifications for voice-only question-answering?
    \item Which effects do prosody modification techniques have on
      informativeness and perceived naturalness of the response?
    \item What type of answers benefit the most from which prosody modifications?
\end{enumerate}

The paper is organized as follows. Section~\ref{sec:relatedwork} discusses related work. Section~\ref{sec:methodology} describes our proposed methodology to evaluate prosody modifications in voice question answering. Section~\ref{sec:setup} details the crowdsourcing experimental setup and Section~\ref{sec:results} discusses the obtained results. Finally, Section~\ref{sec:conclusions} concludes the work by summarizing the answers to the research questions and drawing possible directions for future work.

\section{Related Work}%
\label{sec:relatedwork}

Most of the related work on question answering systems with speech interfaces
focuses on the problem of spoken language recognition and understanding
of voice-based questions~\cite{whittaker2006factoid,mishra2010qme,
rosso2012voice,
kumar2017knowledge}.
The scope of our work is to understand better how to \emph{present} answers when delivered via the audio-channel.

Presenting search results over an audio-channel introduces cognitive challenges for the users because audio is a transient and temporal medium and can blend with environmental sounds~\cite{lai2006speech}.
Thus, a user's short term memory is required to listen and manipulate the presented information to extract the answer from the search results.
However, presenting search results over an audio-only channel can take advantage of the audio signal itself by manipulating the speech synthesis~\cite{arons1997speechskimmer}.
These audio manipulations could similar support the graphical interface provides (e.g., highlighting answers) for users to identify the answer.

In contrast to traditional desktop search, there are no commonly agreed task and evaluation guidelines for assessing
audio modifications for voice-only question answering.
In~\cite{filippova2015sentence} the authors used a crowdsourcing setup to evaluate
sentence compression techniques in terms of readability and informativeness
using human raters.
In contrast to that work, we propose an evaluation setup where human raters are asked to
\emph{listen} and assess the voice answers across \emph{multiple} dimensions,
as well as to \emph{extract} the key answer part, which we later check for correctness.

The audio modifications presented in this paper alter the \emph{prosody} of the spoken answer (i.e., the patterns of stress and intonation in speech).
Speech prosody is one of the major quality dimensions of synthetic speech alongside voice naturalness, fluency, and intelligibility~\cite{hinterleitner2013intelligibility}. It refers to the suprasegmental characteristics of speech such as tune and rhythm. Acoustically, prosody manifests itself in pitch, duration, intensity, and spectral tilt of speech.
Prosody has an essential cognitive role in speech perception~\cite{sanderman1997prosodic}.
Sentence stress seems to ease comprehension of stressed words and has been shown to lower reaction time independent of a word's syntactic function~\cite{cutler1977sentencestress}.
Human listeners attend to those word onsets they are least able to predict~\cite{astheimer2011predictability} and high activation levels allow extra cognitive resources to be allocated for processing these words by the listeners~\cite{cole2010prosodicprominence}.
At signal-level, low-probability regions of pitch and energy trajectories show a strong correlation with the perception of stress, providing further evidence of the connection between attention and unpredictability~\cite{kakouros2016perception}.
Simultaneously, pauses in speech convey information about intonational boundaries~\cite{pannekamp2005prosody} and changes in pausing can alter syntactic parsing of a sentence~\cite{bailey2003disfluencies}. However, interruptions also have a role in comprehension. Filler words, pauses, or even artificial tones have been reported to improve the human word recognition~\cite{corley2011temporaldelay}. This kind of inherent delays and filler words are frequent in spontaneous speech but typically omitted in synthetic speech.

%% file: 02-methodology.tex
\section{Methodology}\label{sec:methodology}

Assume that for a user's \emph{question} we have an \emph{answer sentence}
where we identify the \emph{answer key} (key answer part) with the help of some algorithm.
Table~\ref{tab:example} provides an example of such a tuple.
\begin{table}[tph]
\centering
\small
\caption{Example question and the answer sentence from the SQuAD dataset~\cite{rajpurkar2016squad}.\label{tab:example}}
\begin{tabular}{p{2.5cm}p{5cm}}
\toprule
{\em Question} & Which guitarist inspired Queen?\\
\midrule
{\em Answer Sentence} & Queen drew artistic influence from British rock acts of the 60s [\ldots]
in addition to American guitarist {\bf Jimi Hendrix}, with Mercury also inspired by the gospel singer Aretha Franklin.
\\
\midrule
{\em Answer Key} & Jimi Hendrix \\
\bottomrule
\end{tabular}
\end{table}
This pattern is used by commercial search engines, e.g.,
Google's featured snippets\footnote{\url{https://blog.google/products/search/reintroduction-googles-featured-snippets}}
or Bing Distill answers~\cite{mitra2016proposal}, where
the most important parts of the answer are highlighted or called-out separately.
Additionally, there are datasets available for researchers to study text-based QA,
such as
MS MARCO~\cite{ms-marco}
or the Stanford Question-Answering Dataset~(SQuAD)~\cite{rajpurkar2016squad},
which we use in this paper.

When providing an answer to a user's query on display or screen, it is possible to use visual cues, such as highlighting
or bolding of key answer parts or related terms, which may ease and speed up comprehension of the answer.
In contrast, when serving voice-only answers, one could employ prosody modifications to the speech
to cue the user about the key answer parts.

The problem of identifying key parts is an active area of research in
QA and is beyond the scope of the current work.
Note that, unlike human-curated datasets mentioned above,
the quality of the automatically extracted answer keys may not be high enough for them to be surfaced as stand-alone answers in a production system.
Without the context of the entire sentence, the risk of misleading the user by a low-quality short answer is high.
This potential risk motivates our work on how to \emph{emphasize}
the key part of the answer when presented via voice.

\begin{figure}[tbp]
\centering
\includegraphics[width=0.9\columnwidth]{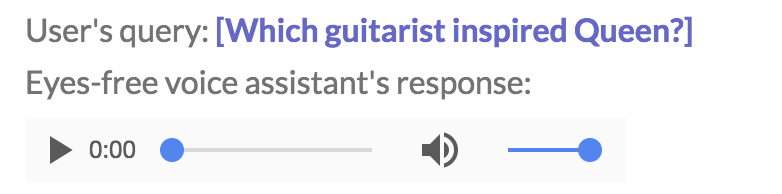}
\caption{Question and audio response as presented to crowd workers.\label{fig:cf-example}}
\end{figure}

We propose to ask \emph{crowd workers} to evaluate
the prosody modifications. %
Given a \emph{question} and verbalization of a corresponding
\emph{answer sentence} (see Figure~\ref{fig:cf-example})
, crowd workers need to
give feedback on the quality of the audio response
as well as identify the phrase in the audio that corresponds to the answer key
(see the Evaluation section below).

We hypothesize that highlighting the key answer part by modifying the prosody
during the audio generation step, makes it easier for the worker
to understand the answer, %
potentially at the expense of naturalness of the audio.

\paragraph*{Prosody modification.}
We perform four different prosody modifications in the Text-to-Speech (TTS) generation:
\begin{itemize}
\item {\bf pause}: inserted before and after the key answer part;
\item {\bf rate}: the speaking rate of the answer key is decreased;
\item {\bf pitch}: the key answer part is spoken in a higher pitch than the rest of the answer sentence;
\item {\bf emphasis}: the answer key is spoken with prominence,
which is typically implemented as a combination of prosody modifications
such as speaking \textbf{rate} and \textbf{pitch}.
\end{itemize}
\noindent%
For instance, the following text input for a TTS engine generates an audio response with {\bf rate} prosodic modification for the example in Table~\ref{tab:example}:
\begin{sloppypar}
\texttt{%
<speak>Queen drew artistic influence from British rock acts of the 60s [\ldots] in addition to American guitarist <prosody rate="slow">Jimi Hendrix</prosody> with Mercury also inspired by the gospel singer Aretha Franklin.</speak>}
\end{sloppypar}
The TTS generated audio with no intervention is used as a baseline against the corresponding TTS system.

\paragraph*{Collection of judgments.}
The judgments were collected using
the Figure Eight crowdsourcing platform (\url{http://www.figure-eight.com}).\footnote{%
All experiments were performed under Ethics Application BSEH 10-14 at RMIT University.}
Figure~\ref{fig:cf-example} shows a question-response pair as presented in the crowdsourcing interface.
The platform randomly assigned tasks to workers, restricting to residents of English-speaking countries.

\paragraph*{Evaluation.}
We study the following four explicit dimensions to evaluate the utility of highlighting via prosody modifications
and naturalness of the audio response:
\emph{informativeness} (how satisfactorily the audio-response answers the user's question on the scale of 0 to 4),
\emph{elocution} (whether the words in the full answer sentence were pronounced correctly, 0 to 2),
presence of unwarranted \emph{interruptions} (0 or 1),
appropriateness of the audio \emph{length} (-1 to 1). 
These dimensions are based on the guidelines for evaluating speech in the Google Assistant,~\cite{GoogleAssistantGuidelines}
and the exact version used for crowdsourcing,
along with the dataset and judgments collected can be accessed at \url{https://github.com/rmit-ir/clef2019-prosody}.
In addition to collecting the aforementioned judgments, we also calculate one objective measure,
the \emph{correctness} of the workers' typed answer key.
To compute correctness, we compare the answer key typed by the worker
against the given short answer from the dataset (what we treat as the gold answer key for highlighting).
We convert both into a Metaphone representation~\cite{philips2000double} to account for typos and misheard words, and then compute the difference using the Ratcliff-Obershelp algorithm~\cite{ratcliff1988pattern}. The \emph{correctness} value ranges from 0 to 1.

\paragraph*{Quality control.}
To detect whether a worker is reliable, we use two different types of \emph{test questions}:
\begin{inparaenum}[(i)]
    \item we ask the worker to type in the short answer after listening to the full audio and
    then compare the provided short answer against the ground-truth, and
    \item we include questions that are off-topic and do not contain an answer%
\end{inparaenum}.
In the first case, we filter out workers who achieve answer \emph{correctness} below $0.5$ while in the second case we expect the worker to give the lowest rating on the \emph{informativeness} scale.

%% file: 03-setup.tex
\section{Experimental Setup}
\label{sec:setup}

In our study, we use question/answer pairs
from the widely used SQuAD question-answering dataset~\cite{rajpurkar2016squad}.
The SQuAD dataset consists of crowdsourced questions related to a set of Wikipedia articles.
Each Wikipedia article is split into paragraphs, and each paragraph has
a set of questions.
Answers to those questions correspond to text segments of the corresponding paragraph.
In our experiments, the paragraph is fed to a TTS to generate the audio response
and the ground truth answers are used to highlight the key part of it.

In particular, for our set of experiments, we used the first 300 Wikipedia
articles and their corresponding question/audio pairs. We further split these
300 articles into four groups of 75 question/audio pairs (one group per
modification: pause, rate, pitch, and emphasis). Choosing different questions
for each prosody modification reduces the chance that each crowd worker is
exposed to the same question multiple times, something that we cannot easily
enforce in the crowdsourcing platform we used.
We then generated the audio of the
baseline and modified versions of the answer sentence.
Three crowd workers rated each of the resulting question/audio pairs.

We use two different TTS platforms in our experiments:
IBM Watson~\cite{sorin2017semi}
and Google Wavenet~\cite{van2016wavenet}, accessed at
\url{https://ibm.com/watson/services/text-to-speech} and \url{https://cloud.google.com/text-to-speech} respectively.
The settings are summarized in Table~\ref{tab:ttssettings}.\footnote{The
\texttt{emphasis} feature is currently only available in the Google TTS and
the implementation details are not specified in the SSML standard~\cite{SSMLstandard}
nor the documentation.}
Note that these settings are chosen ad-hoc based on the subjective
perception of the authors and test runs. The perceived size of the effect depends on
the TTS engine and voice used, as well as on the sentence being modified.
We leave the optimization of the level of prosodic modifications for future work.
\begin{table}
  \centering
    \caption{Prosody modification settings: \texttt{strength} parameter of
    the \texttt{<break>} SSML tag, \texttt{rate} / \texttt{pitch} parameters of \texttt{<prosody>},
    and \texttt{level} parameter of \texttt{<emphasis>}.}
    \begin{tabular}{p{1cm}@{~}l@{~}l@{~~}l@{~~}l@{~~}l}    \toprule
      TTS \mbox{engine} & Voice & \textbf{pause} & \textbf{rate} & \textbf{pitch} & \textbf{emphasis} \\
    \midrule
      IBM & Lisa & \texttt{strong} & \texttt{x-slow} & \texttt{x-high} & n/a \\
      Google & \mbox{Wavenet-F}& \texttt{strong} & \texttt{slow} & \texttt{+2st} & \texttt{strong} \\
    \bottomrule
  \end{tabular}
  \label{tab:ttssettings}
\end{table}

After removing judgments used for quality control, we have \num{1454} rows of judgments for the IBM engine from 99 workers for 450 question-audio pairs (75 for each of the three modifications (pause, rate, pitch), plus an equal number of baseline pairs); \num{1820} rows of judgments for the Google TTS engine from 85 workers for 600 question-audio pairs (75 for each of the four modifications, plus an equal number of baseline pairs). 

The agreement between crowd workers is rather low when measured
by Krippendorff's alpha %
as seen in Table~\ref{tab:agreement}, especially for
\emph{elocution} and \emph{interruption} scores
(e.g., only for \textbf{pauses} modification did we have meaningful agreement
for \emph{interruption} score).
For \emph{informativeness} and \emph{length} the scores are low, but are
comparable with similar crowdsourcing judgment collections~\cite{chuklin2016incorporating}.
When it comes to majority agreement, however, it was substantially high
across all dimensions/modifications/voices, meaning that two out of three
workers selected the same rating label for almost all items.

\begin{table*}[tbp]
    \centering
    \caption{Krippendorff's $\alpha$ and ratio of items where majority of the raters agree on the rating; ranges given for different modifications.}
        \begin{tabular}{lrrrr}
        \toprule
                score       &       $\alpha$ (IBM)   &  $\alpha$ (Google)   & majority\ (IBM)    & majority\ (Google) \\
        \midrule
    \emph{informativeness} & \numrange[range-phrase = \dots]{0.27}{0.31} & \numrange[range-phrase = \dots]{0.06}{0.22} &  \numrange[range-phrase = \dots]{0.84}{0.87}  & \numrange[range-phrase = \dots]{0.79}{0.89} \\
    \emph{elocution}    & \numrange[range-phrase = \dots]{0.15}{0.27} & \numrange[range-phrase = \dots]{-0.04}{0.08} & \numrange[range-phrase = \dots]{0.87}{0.97} & \numrange[range-phrase = \dots]{0.99}{1.00} \\
    \emph{interruption} & \numrange[range-phrase = \dots]{0.00}{0.08} & \numrange[range-phrase = \dots]{-0.01}{0.12} & \numrange[range-phrase = \dots]{0.99}{1.00} & \numrange[range-phrase = \dots]{1.00}{1.00} \\
    \emph{length}  & \numrange[range-phrase = \dots]{0.27}{0.37} & \numrange[range-phrase = \dots]{0.17}{0.43}& \numrange[range-phrase = \dots]{0.97}{0.99} & \numrange[range-phrase = \dots]{0.99}{1.00} \\
        \bottomrule
    \end{tabular}
    \label{tab:agreement}
\end{table*}

Judgments are treated as Likert scale and, in case of \emph{length},
the absolute value is taken, effectively making it binary (``OK'' vs. ``too short or too long'').
We use the median to aggregate judgments per item.
Given that the ratings are not on the interval scale,
we use the Wilcoxon signed-rank test~\cite{wilcoxon1945individual}
on a per-item level to report statistical significance.
We use $^*$ ($^{**}$) to indicate statistical significance with $p<0.05$ ($p<0.01$ respectively).
Equivalent results were obtained when the t-test and/or average instead of median was used.

%% file: 04-results.tex
\section{Results and Discussion}
\label{sec:results}

This section is structured as follows. First we discuss the results of the audio modifications against the baseline audio which had no modifications for all the different question answering pairs (\S\ref{subsec:generalresults}). Then we investigate the differences for question answering pairs where the key answer part was short (\S\ref{subsec:Answer Length}), where the answer sentence was long (\S\ref{subsec:Support Sentence Length}), and finally we analyze the offset of the answer key from the end of answer sentence (\S\ref{subsec:Answer Offset From the End of Support Sentence}).

\subsection{General Results}
\label{subsec:generalresults}

Table~\ref{tab:results} shows the difference in terms of the judgments
obtained for the proposed prosody modifications, using the two TTS systems,
against the baseline for each corresponding TTS system, without prosody modification.
We only report the difference in the score given by the raters (absolute numbers)
to avoid a direct comparison between two commercial systems.
Note also that the results are not comparable across two systems because the prosody
modifications with the same name have a noticeably different effect on them.

\input{tbl-results}

The main pattern that emerges from the data is an increase in informativeness and correctness, and a decrease in speech quality through naturalness, as captured by \emph{elocution} or \emph{interruption} ratings.
Interestingly, only \textbf{rate} modification was deemed to significantly hurt elocution and no significant \emph{length} change were reported.
As expected, workers identified more unexpected \emph{interruptions} when \textbf{pauses} are used to highlight the answer keys in both TTS engines.

Note, that there were also \emph{interruptions} reported for other modifications in the Google TTS engine. This is due to the peculiarity of that engine, which always adds sentence breaks---and therefore small pauses---around \texttt{<prosody>} and \texttt{<emphasis>} tags. We expect that once that issue is resolved, no interruptions will be reported.

We also observe that our prosody modifications either
improve or leave the \emph{correctness} score unchanged,
and most of them---although not all---are perceived by workers
as more useful for the job of identifying the answer (\emph{informativeness}).

\smallskip\noindent%
Next, we look at whether particular prosody modifications are especially
effective (or not) on specific slices of the data.
We do it by splitting the data by
the median length of the key answer part
and its median position in the answer sentence, thereby roughly halving
the data into two similarly-sized slices in each case.
Table~\ref{tab:resultslen} demonstrates how different prosody modifications
strategies perform depending on the length of the short answer.
Table~\ref{tab:resultssentencelen} shows the split by the \emph{sentence} length.
Finally, Table~\ref{tab:resultsoffset} shows how results change depending on the
short answer offset from the end of the support sentence.
Both offset and length were measured in number of words, but similar results
were obtained when measuring the number of characters.

\subsection{Answer Key Length}
\label{subsec:Answer Length}

\input{tbl-results-length}

As we can see from Table~\ref{tab:resultslen}, \textbf{rate} change in the IBM
engine is perceived to increase \emph{informativeness} for shorter keys,
while \textbf{pitch} appears to work better for the longer ones.
For the Google engine, we observe a similar pattern for \textbf{pauses}
which are very effective for emphasizing short answer keys.
However, pauses has a negative effect when the answer keys are longer,
as measured by both
subjective (\emph{informativeness}) and objective (\emph{correctness}) scores.
On the other hand, \textbf{emphasis}, has a positive effect on both slices.
We also see that the effect mentioned above of unwanted interruptions in the Google engine is not noticeable when longer phrases are emphasized.
Note that the implementations differ between TTS engines, especially
for \textbf{rate} and \textbf{pitch} settings where we were unable to exactly
match the perceived strength of the modifications in two engines.\footnote{%
We selected maximum strength for \textbf{rate} and \textbf{pitch} in the IBM
TTS engine, but the settings with the same labels in the Google TTS
were subjectively too strong and unnatural, so we chose to tune them down.}

\subsection{Answer Sentence Length}
\label{subsec:Support Sentence Length}

\input{tbl-results-sentence-length}

Table~\ref{tab:resultssentencelen} shows similar breakdown, in this case
by the length of the answer sentence.

We can see that long answer sentences
are naturally hard to digest which leads to lack of statistical significance
in the \emph{informativeness} and \emph{correctness} scores.
For the short sentences we see a picture similar to the overall results
presented in~Table~\ref{tab:results}, but the \textbf{rate} modification
of the IBM engine now statistically significantly improves \emph{informativeness}
and \emph{correctness} ($p < 0.05$).

\subsection{Answer Key Offset From the End of the Sentence}
\label{subsec:Answer Offset From the End of Support Sentence}

\input{tbl-results-offset}

Finally, Table~\ref{tab:resultsoffset} shows a breakdown by the offset
to the end of the answer sentence.
Intuitively, when the answer key is closer to the end of the audio, it should be easier to understand. %
For these answers we observed only weak significant improvement
of the \textbf{rate} modification for the IBM engine
as measured by \emph{correctness} and, to a certain extent,
by \emph{informativeness}. %
On the same slice the Google engine benefits from both \textbf{rate}
and \textbf{emphasis} modifications as measured by both \emph{informativeness}
and \emph{correctness}.

\begin{figure}
    \centering
    \includegraphics[width=0.8\linewidth]{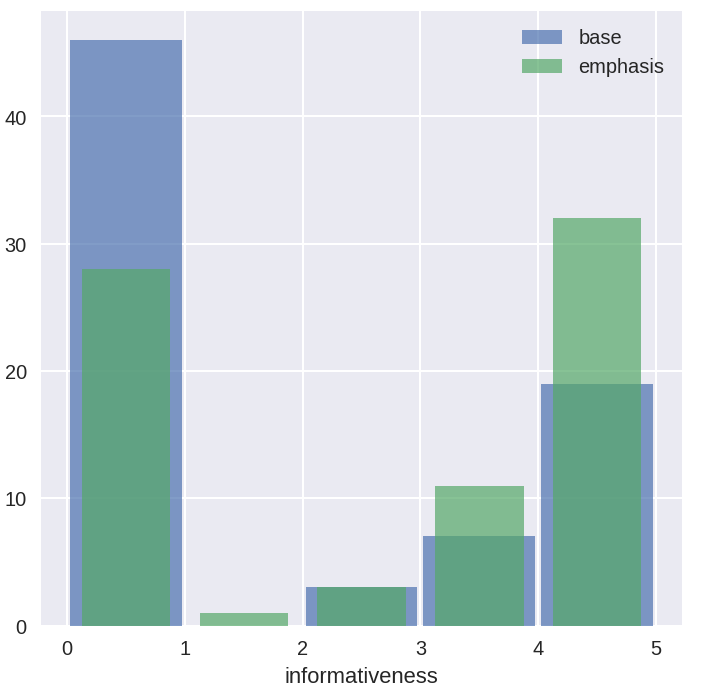}
    \includegraphics[width=0.8\linewidth]{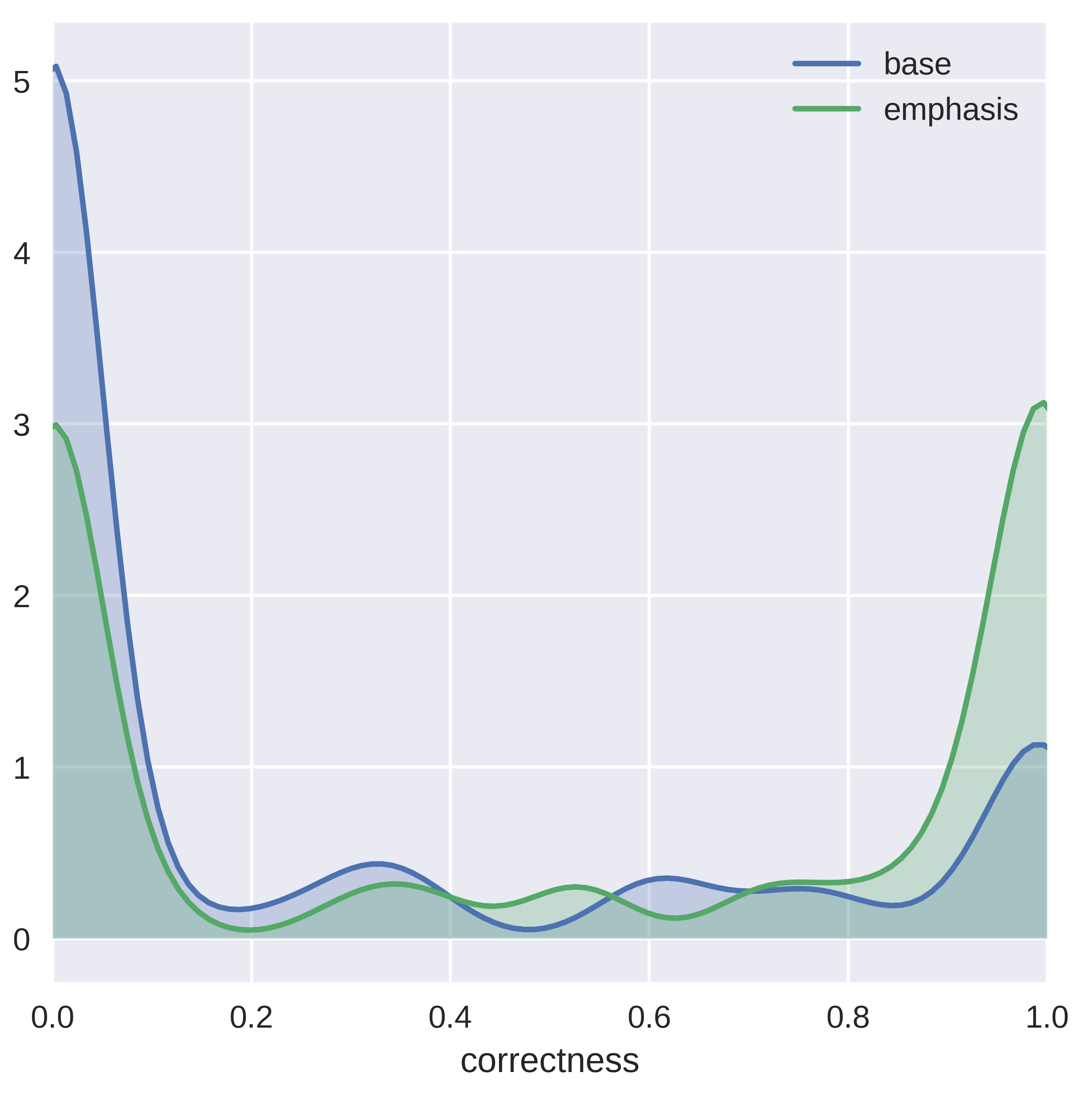}
    \caption{Distributions of the \emph{informativeness} (above) and
\emph{correctness} (below) of the audio with emphasis (red narrow bars)
vs.\ the baseline (blue wider bars). Higher scores are better.}
    \label{fig:emphasis}
\end{figure}
\medskip
\noindent%
Overall, the \textbf{emphasis} modification in the Google engine obtains the highest
(and statistically significant) gain in terms of
\emph{informativeness} and \emph{correctness} (see Figure~\ref{fig:emphasis}
for a visual representation of the distribution shift).
That means that the combination of prosody modifications developed by
the engineers of this feature outperforms modifications of just one dimension.
Further exploration of how to combine these modifications needs to be done.

%% file: tbl-results.tex
\begin{table*}
    \centering
    \caption{Difference relative to the baseline (no modifications); absolute value.
    The higher the better for \emph{informativeness}, \emph{correctness}, and \emph{elocution} ($\uparrow$);
    the lower the better for \emph{interruption} and \emph{length} ($\downarrow$).
    }
\begin{tabular}{llllll}
\toprule
 & \emph{informativeness$\uparrow$} & \emph{correctness$\uparrow$} & \emph{elocution$\uparrow$} & \emph{interruption$\downarrow$} & \emph{length$\downarrow$} \\
\midrule
IBM &&&&& \\
\textbf{pauses} &                $-0.21$ &            $+0.04$ &          $-0.03$ &        $+0.37^{**}$ &       $+0.08$ \\
\textbf{rate}   &                $+0.26$ &            $+0.02$ &     $-0.24^{**}$ &             $+0.03$ &       $+0.03$ \\
\textbf{pitch}  &                $+0.02$ &            $-0.03$ &          $-0.11$ &             $+0.01$ &       $-0.03$ \\
\midrule
Google &&&&& \\
\textbf{pauses}   &                $+0.21$ &            $+0.09$ &          $-0.04$ &        $+0.15^{**}$ &       $+0.00$ \\
\textbf{rate}     &                $+0.22$ &            $+0.07$ &     $-0.18^{**}$ &        $+0.18^{**}$ &       $+0.03$ \\
\textbf{pitch}    &                $-0.03$ &            $+0.08$ &          $+0.08$ &        $+0.13^{**}$ &       $+0.07$ \\
\textbf{emphasis} &           $+0.87^{**}$ &       $+0.28^{**}$ &          $-0.07$ &        $+0.13^{**}$ &       $-0.07$ \\
\bottomrule
\end{tabular}
\label{tab:results}
\end{table*}

%% file: tbl-results-length.tex
\begin{table*}
    \centering
    \caption{Difference relative to the baseline (no modifications); absolute value.
Short answer keys are equal in length or shorter than the median ones (two words), while long is the rest.}
\begin{tabular}{llllll}
\toprule
 & \emph{informativeness$\uparrow$} & \emph{correctness$\uparrow$} & \emph{elocution$\uparrow$} & \emph{interruption$\downarrow$} & \emph{length$\downarrow$} \\
\midrule
\multicolumn{6}{l}{IBM (short answer keys)} \\
\textbf{pauses} &                $-0.12$ &            $+0.09$ &          $-0.05$ &        $+0.37^{**}$ &       $+0.16$ \\
\textbf{rate}   &                $+0.48$ &            $+0.07$ &          $-0.16$ &             $+0.00$ &       $-0.10$ \\
\textbf{pitch}  &                $-0.14$ &            $-0.08$ &          $-0.06$ &             $+0.02$ &       $-0.05$ \\
\midrule
\multicolumn{6}{l}{IBM (long answer keys)} \\
\textbf{pauses} &                $-0.33$ &            $-0.03$ &          $+0.00$ &        $+0.36^{**}$ &       $-0.03$ \\
\textbf{rate}   &                $-0.02$ &            $-0.05$ &        $-0.35^*$ &             $+0.06$ &       $+0.20$ \\
\textbf{pitch}  &                $+0.24$ &            $+0.05$ &          $-0.18$ &             $+0.00$ &       $-0.02$ \\
\midrule
\multicolumn{6}{l}{Google (short answer keys)} \\
\textbf{pauses}   &           $+1.09^{**}$ &       $+0.27^{**}$ &          $-0.07$ &        $+0.19^{**}$ &       $-0.05$ \\
\textbf{rate}     &                $+0.11$ &            $+0.08$ &     $-0.19^{**}$ &        $+0.25^{**}$ &       $+0.03$ \\
\textbf{pitch}    &                $-0.05$ &            $+0.10$ &          $+0.02$ &        $+0.16^{**}$ &       $+0.07$ \\
\textbf{emphasis} &                $+0.54$ &          $+0.21^*$ &          $-0.12$ &        $+0.17^{**}$ &       $-0.05$ \\
\midrule
\multicolumn{6}{l}{Google (long answer keys)} \\
\textbf{pauses}   &              $-0.97^*$ &          $-0.16^*$ &          $+0.00$ &             $+0.09$ &       $+0.06$ \\
\textbf{rate}     &                $+0.38$ &            $+0.06$ &        $-0.16^*$ &             $+0.09$ &       $+0.03$ \\
\textbf{pitch}    &                $+0.00$ &            $+0.05$ &          $+0.16$ &             $+0.10$ &       $+0.06$ \\
\textbf{emphasis} &           $+1.26^{**}$ &       $+0.36^{**}$ &          $+0.00$ &             $+0.09$ &       $-0.09$ \\
\bottomrule
\end{tabular}
\label{tab:resultslen}
\end{table*}

%% file: tbl-results-sentence-length.tex
\begin{table*}
    \centering
    \caption{Difference between the baseline (without modifications) and the various prosody modification and various answer sentence lengths.
Short sentences are equal in length or shorter than the median sentence (measured by the number of words), long is the rest.}
    \begin{tabular}{llllll}
\toprule
 & \emph{informativeness$\uparrow$} & \emph{correctness$\uparrow$} & \emph{elocution$\uparrow$} & \emph{interruption$\downarrow$} & \emph{length$\downarrow$} \\
\midrule
\multicolumn{6}{l}{IBM (short sentences)} \\
\textbf{pauses}   &                $-0.47$ &            $-0.05$ &          $-0.01$ &        $+0.35^{**}$ &     $+0.24^*$ \\
\textbf{rate}     &              $+0.98^*$ &          $+0.17^*$ &     $-0.35^{**}$ &             $+0.03$ &     $-0.19^*$ \\
\textbf{pitch}    &                $+0.04$ &            $-0.01$ &          $-0.13$ &             $+0.02$ &       $+0.04$ \\
\midrule
\multicolumn{6}{l}{IBM (long sentences)} \\
\textbf{pauses}   &                $+0.10$ &            $+0.14$ &          $-0.04$ &        $+0.39^{**}$ &       $-0.10$ \\
\textbf{rate}     &                $-0.23$ &            $-0.09$ &          $-0.17$ &             $+0.02$ &       $+0.18$ \\
\textbf{pitch}    &                $-0.02$ &            $-0.06$ &          $-0.06$ &             $+0.00$ &       $-0.19$ \\
\midrule
\multicolumn{6}{l}{Google (short sentences)} \\
\textbf{pauses}   &                $+0.32$ &            $+0.06$ &          $+0.03$ &           $+0.12^*$ &       $+0.03$ \\
\textbf{rate}     &                $+0.48$ &            $+0.17$ &     $-0.30^{**}$ &        $+0.33^{**}$ &       $+0.04$ \\
\textbf{pitch}    &                $+0.00$ &            $+0.12$ &          $-0.02$ &           $+0.11^*$ &       $+0.09$ \\
\textbf{emphasis} &           $+0.96^{**}$ &       $+0.32^{**}$ &          $-0.09$ &        $+0.13^{**}$ &       $-0.09$ \\
\midrule
\multicolumn{6}{l}{Google (long sentences)} \\
\textbf{pauses}   &                $+0.12$ &            $+0.11$ &          $-0.10$ &        $+0.17^{**}$ &       $-0.02$ \\
\textbf{rate}     &                $+0.08$ &            $+0.02$ &        $-0.11^*$ &           $+0.10^*$ &       $+0.03$ \\
\textbf{pitch}    &                $-0.07$ &            $+0.00$ &     $+0.24^{**}$ &           $+0.17^*$ &       $+0.03$ \\
\textbf{emphasis} &                $+0.64$ &            $+0.19$ &          $+0.00$ &             $+0.14$ &       $+0.00$ \\
\bottomrule
\end{tabular}
\label{tab:resultssentencelen}
\end{table*}

%% file: tbl-results-offset.tex
\begin{table*}
    \centering
    \caption{Difference relative to the baseline (no modifications); absolute value.
``Easy'' answers have answer key positioned at equal or less than median offset from the sentence end, while ``hard'' are the rest.}
\begin{tabular}{llllll}
\toprule
 & \emph{informativeness$\uparrow$} & \emph{correctness$\uparrow$} & \emph{elocution$\uparrow$} & \emph{interruption$\downarrow$} & \emph{length$\downarrow$} \\
\midrule
IBM (easy)  &&&&& \\
\textbf{pauses} &                $+0.02$ &            $+0.08$ &          $-0.05$ &        $+0.31^{**}$ &       $+0.06$ \\
\textbf{rate}   &                $+0.60$ &          $+0.12^*$ &        $-0.26^*$ &             $+0.06$ &       $-0.03$ \\
\textbf{pitch}  &                $-0.20$ &            $-0.08$ &          $-0.10$ &             $+0.02$ &       $+0.01$ \\
\midrule
IBM (hard) &&&&& \\
\textbf{pauses} &                $-0.53$ &            $-0.02$ &          $+0.00$ &        $+0.45^{**}$ &       $+0.11$ \\
\textbf{rate}   &                $-0.02$ &            $-0.07$ &        $-0.23^*$ &             $+0.00$ &       $+0.07$ \\
\textbf{pitch}  &                $+0.28$ &            $+0.04$ &          $-0.12$ &             $+0.00$ &       $-0.09$ \\
\midrule
Google (easy)  &&&&& \\
\textbf{pauses}   &                $+0.61$ &          $+0.15^*$ &          $-0.07$ &        $+0.15^{**}$ &       $-0.04$ \\
\textbf{rate}     &           $+1.01^{**}$ &       $+0.24^{**}$ &        $-0.15^*$ &        $+0.19^{**}$ &       $-0.07$ \\
\textbf{pitch}    &                $+0.09$ &            $+0.08$ &          $+0.07$ &           $+0.14^*$ &       $+0.12$ \\
\textbf{emphasis} &           $+1.23^{**}$ &       $+0.35^{**}$ &          $-0.16$ &           $+0.13^*$ &       $-0.03$ \\
\midrule
Google (hard)  &&&&& \\
\textbf{pauses}   &                $-0.41$ &            $-0.02$ &          $+0.00$ &           $+0.14^*$ &       $+0.07$ \\
\textbf{rate}     &                $-0.49$ &            $-0.07$ &     $-0.20^{**}$ &        $+0.17^{**}$ &       $+0.13$ \\
\textbf{pitch}    &                $-0.19$ &            $+0.07$ &          $+0.09$ &           $+0.12^*$ &       $+0.00$ \\
\textbf{emphasis} &                $+0.61$ &          $+0.23^*$ &          $+0.00$ &           $+0.14^*$ &       $-0.09$ \\
\bottomrule
\end{tabular}
\label{tab:resultsoffset}
\end{table*}

%% file: 05-conclusions.tex
\section{Conclusions}\label{sec:conclusions}
We investigate how
prosody modifications can help users to identify answers from audio responses
in a question answering setting.
To answer our first research question (\textbf{RQ1}) %
we conclude that, yes,
the proposed crowdsourcing setup is viable and gives an actionable breakdown of quality dimensions.
To our knowledge, this is the first experiment that validates
the use of a crowdsourcing methodology to analyze prosody modification
in voice-only question answering.

Answering our second research question (\textbf{RQ2}), we show that emphasizing the answer---via lowering speaking rate and simultaneously increasing pitch---provides subjectively more informative responses and makes workers more effective in identifying the answers, at the expense of the naturalness in the audio (interruptions), which is an artifact of a particular implementation.

Finally, to answer 
the last research question (\textbf{RQ3})
we conclude that the sentences where the key part is found closer to the end,
are more likely to get benefit from audio transformations for highlighting the answer key.
In general, however, different highlighting techniques work best on different answer slices.

The near future work includes further studies to find the optimal combination
of prosody modification to highlight answers in a given audio response depending
on the different answer features (and possibly on the user features).
Another open question for future work is to better understand how modifying the prosody
impacts the users' comprehension and satisfaction in a more general context,
such as when users are not asked to extract answers.